\documentclass[lettersize,journal]{IEEEtran}

\usepackage[utf8]{inputenc} 
\usepackage[T1]{fontenc}    
\usepackage{hyperref}       
\usepackage{url}            
\usepackage{booktabs}       
\usepackage{amsfonts}       
\usepackage{nicefrac}       
\usepackage{microtype}      
\usepackage{graphicx}
\usepackage{csquotes}
\usepackage[caption=false,font=footnotesize]{subfig}
\usepackage{adjustbox}
\usepackage{amsmath}
\usepackage{adjustbox}
\usepackage[dvipsnames,table]{xcolor}

\usepackage[linesnumbered,ruled,vlined]{algorithm2e}

\title{Perception Without Vision for Trajectory Prediction: \\ Ego Vehicle Dynamics as Scene Representation for \\ Efficient Active Learning in Autonomous Driving}

\author{Ross~Greer~\IEEEmembership{Member,~IEEE},
        Mohan~M.~Trivedi~\IEEEmembership{Life Fellow,~IEEE}
\thanks{All authors are with the Laboratory for Intelligent and Safe Automobiles, Department
of Electrical and Computer Engineering, University of California San Diego, La Jolla, CA, 92092 USA. 
Corresponding Author E-mail: regreer@ucsd.edu.}
}

\begin{document}

\maketitle

\begin{abstract}

This study investigates the use of trajectory and dynamic state information for efficient data curation in autonomous driving machine learning tasks. We propose methods for clustering trajectory-states and sampling strategies in an active learning framework, aiming to reduce annotation and data costs while maintaining model performance. Our approach leverages trajectory information to guide data selection, promoting diversity in the training data. We demonstrate the effectiveness of our methods on the trajectory prediction task using the nuScenes dataset, showing consistent performance gains over random sampling across different data pool sizes, and even reaching sub-baseline displacement errors at just 50\% of the data cost. Our results suggest that sampling typical data initially helps overcome the "cold start problem," while introducing novelty becomes more beneficial as the training pool size increases. By integrating trajectory-state-informed active learning, we demonstrate that more efficient and robust autonomous driving systems are possible and practical using low-cost data curation strategies.
\end{abstract}

\begin{IEEEkeywords}
autonomous driving, trajectory prediction, active learning, data curation, clustering
\end{IEEEkeywords}

\section{Introduction}

The accurate prediction of the trajectories of agents in the observed environment is paramount to the safe path planning of autonomous systems. Whether the agents are observed from infrastructure, the ego vehicle, or some combination of modalities, forecasting where other vehicles and pedestrians helps intelligent systems (human and machine alike) to make their own control decisions. 

Machine learning has provided a means for trajectory prediction of traffic agents using rasterized bird's-eye-view maps, contextual scene information, and social dynamics \cite{cui2019multimodal, messaoud2021trajectory}. Road infrastructure \cite{kim2022diverse, greer2021trajectory}, agent occupancy \cite{deo2018convolutional}, and navigation goals \cite{deo2020trajectory} largely determine where and how a vehicle will move through the environment. However, collection and especially annotation of data for such systems can be costly. Methods in trajectory prediction and planning rely on the ability to perceive road infrastructure and agents; for example, in methods which use a bird's-eye-view map of the scene to predict a trajectory, the data must include accurate annotations of the position of scene agents, lane markings, and intersections. While the trajectory itself can be quickly collected from onboard positioning sensors, the annotation of the surrounding scene which informs the driving decision-making is a costly effort \cite{zimmer20193d, greer2024and}. In this research, we consider the utility of trajectory data as the basis of acquisition functions for the purposes of active, semi-, or self-supervised learning \cite{ruckin2024semi, almin2023navya3dseg, ghita2024activeanno3d}; in other words, how might information on a vehicle's positioning help us to curate data for efficient machine learning using minimal annotation budgets?

\begin{figure}
    \centering
    \includegraphics[width=.5\textwidth]{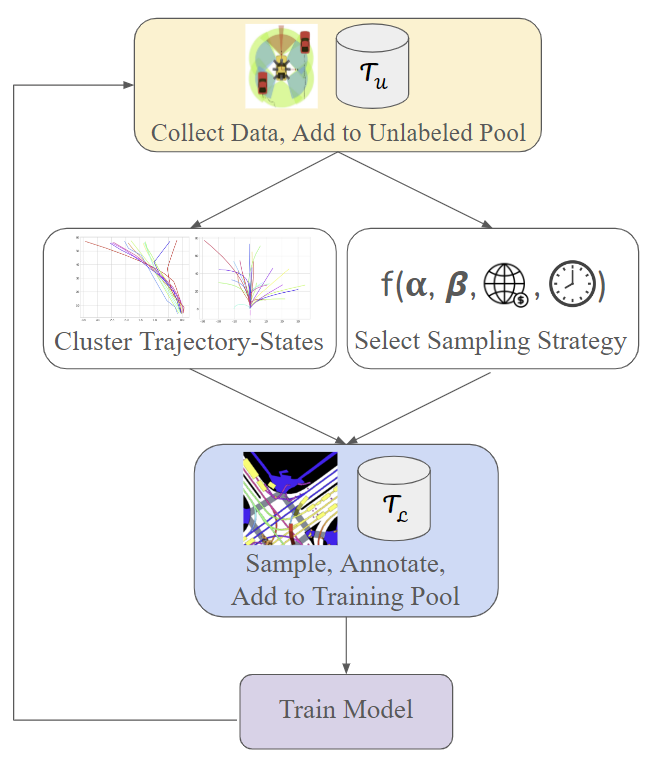}
    \caption{In supervised learning for tasks such as trajectory prediction, data is collected (yellow), annotated and added to a training pool (blue), and then a model is trained (purple). When more data is collected than can be afforded by an annotation or computational budget, intelligent sampling using active learning (white) may provide solutions which maintain model performance at reduced data cost. We contribute algorithms for clustering of trajectory-states and sampling strategies which are model-agnostic, providing a benefit of active learning based only on the current training data and without requiring computation of uncertainty from the partially-trained model.}
    \label{fig:enter-label}
\end{figure}

To help illustrate this idea, consider a situation where you are a passenger in a vehicle driving with modern advanced driver assistance functionality, and perhaps you are tired and decide to close your eyes. You may experience a variety of kinematic cues even without vision; you may feel the car come to a stop, and after a few moments (or perhaps a bit longer), you feel the car turn to the left, then continue smoothly. Even though you have no vision of the environment, there are many pieces of information which you can already gleam from these dynamics alone. First, you came to a stop - this does not happen without a reason. Perhaps you approached a stop sign, a red traffic light, or a person crossing the road. You then waited for a bit (presumably, enough to come to a complete stop and wait until safe to proceed, or the light turns green, or the person finishes crossing). Then, you made a left turn, meaning that you were likely at some kind of intersection, and depending on your wait, possibly with other agents. In any of the above cases, from the trajectory alone, you would be able to reasonably infer that you are not cruising on the freeway - and with enough examples like this, you may be able to recognize patterns in the dynamics which relate to the outside scene, all without observing the outside scene!

In this way, we propose that trajectory information shares mutual information with the visual observation of a scene, and that we can use this trajectory information in an unsupervised manner to inform our data curation process for autonomous driving machine learning tasks, to promote diversity in our data. Having data which covers the input space as thoroughly as possible is critical to robust learning \cite{sener2017active}.

Towards the continued development of such techniques, this research presents methods of curating and integrating further training data for such systems, such that systems can efficiently learn new behavioral patterns and adapt to changes in the open set of real-world driving scenarios. Our contributions are as follows:
\begin{enumerate}
    \item Demonstration of the ability of trajectory and vehicle dynamic state information to be clustered for the purposes of learning acquisition functions, and algorithms for such acquisition in active and continual learning settings, 
    \item Presentation of sampling techniques related to (a) breadth and depth of data clusters and (b) introduction of novelty, 
    \item Discussion of the relevant data features toward an example task of trajectory prediction, with relation to the cost of annotation and benefit to learning systems, and discussion of extension to related tasks of object detection and path planning, 
    \item Empirical analysis of the learning phase transition with respect to novel data, and
    \item Empirical analysis of the effectiveness of novelty-sensitive sampling in an active learning experiment, illustrating the potential of the system to continually learn from intelligently-selected new data.
\end{enumerate}

\section{Related Research}

Machine learning relies on transforming data into separable representations, and to do so effectively, requires training data which approximately covers the variance of data expected to be encountered in deployment in the real-world. To this end, active learning is a method by which data is incrementally annotated and added to a training pool for a machine learning system, selected in a strategic manner for efficiency over an annotation budget. Broadly, these methods are divided into uncertainty-driven methods, which take into account a model's level of confidence in its prediction of an unlabeled datum, and diversity-driven methods, which take into account the relationship of a datum to all other data \cite{yang2015multi, lu2024activead}. Active learning has been useful in supporting a variety of autonomous driving tasks such as vehicle detection, recognition, and tracking \cite{sivaraman2010general, satzoda2015multipart}. 

Hacohen et al. define, derive, and empirically support the existence of an active learning ``phase transition" in model performance with respect to data \textit{typicality} \cite{hacohen2022active}. The term \textit{typicality} is used to describe points in a high-density region of the input space, analogous but opposite to the meaning of \textit{diversity} for such tasks, and without regard for model \textit{certainty}. Hacohen et al. show that on low budgets, sampling typical data is most beneficial, while on high budgets, sampling least typical data is most beneficial. They evaluate their hypothesis on three image classification tasks (CIFAR-10, CIFAR-100, and ImageNet-100). Important to this research, they also remind readers that their work is especially relevant for applications which require ``an expert tagger whose time is expensive", and autonomous driving certainly falls into this category, where companies frequently outsource data annotation to teams of taggers \cite{caesar2020nuscenes}, whose expertise and attention directly influence safety outcomes of algorithms trained on this annotated data \cite{rottmann2023automated, kulkarni2021create}. Their discussion of the importance is not just related to efficiency; when data is within the ``low budget" regime, general active learning methods fail to surpass random sampling! This is referred to as the \textit{cold start problem} \cite{zhu2019addressing}, and may be a consequence of early models being without the critical mass of data to form accurate measurements of its ``uncertainty" of unlabeled points. 

This provides a few implications relevant for tasks in autonomous driving and, more generally, robotics: 
\begin{enumerate}
    \item It is important to identify at what data volume this phase transition occurs. Without awareness of the phase transition, because of the cold start problem, one cannot identify whether to employ active learning, and even then which active learning method to employ. 
    \item Once the phase transition is identified, selecting the right learning strategy will depend on defining a notion of ``typicality" (or, in dual, ``novelty") which is pertinent to the domain, task, and data at hand. 
\end{enumerate}

In our experiments, we present evidence for this phase transition within data systems for an example task of trajectory prediction, and provide a measurement of typicality useful for clustering such data in the domain. 

\section{Novelty-Sensitive Active Learning Algorithm using Trajectories and Dynamic States}

Trajectory and dynamic information \cite{satzoda2014drive} is particularly low-cost to collect and annotate. Assuming a well-calibrated GPS and IMU system, the vehicle is localized and trajectories can be reconstructed in a 2D overhead projection, along with state variables such as velocity, accleration, and heading. This requires virtually no annotation, as opposed to 2D or 3D objects in a scene, which require meticulous annotation. In the next sections, we describe ways that the low-cost information can be leveraged to curate only particular, learning-efficient scenes (which can then be expensively annotated) for an overall reduction in data budget while maintaining performance.

\subsection{Sampling Iteration Parameterization}

We begin with an assumption that it is possible to identify novel data using unsupervised techniques, which we detail in the following sections. We adopt a clustering approach, where any data sufficiently distant from centers of clusters with members in the training pool are considered novel. From this, we consider two parameters which define our sampling mechanism: $\alpha$, representing the proportion of novel data which should be sampled (where $1-\alpha$ is the proportion of training-pool-similar data to be sampled), and $\beta$, the proportion of each cluster allowed to be sampled (in other words, how many instances of a novel concept can be added in a sampling iteration). For a fixed annotation budget, $\alpha$ and $\beta$ can be tuned to manage the breadth of novel clusters sampled and the depth with which a novel cluster is sampled. 

\subsection{Acquisition Function using Trajectory and State Similarity Clustering}

For the purposes of prediction, one formulation of an autonomous driving trajectory involves the combination of 2D ground plane world coordinates that the vehicle will occupy for $n$ seconds sampled at rate $r$, as well as any initial state variables $s$ that describe the agent at the beginning of the prediction period. In the case of nuScenes, for example, data is sampled at 2 Hz for 6 seconds, and $s$ is comprised of the vehicle velocity $v$, accleration $a$, and heading change rate $h$.

We define a measure of similarity between two vehicle trajectory-states $V_i$ and $V_j$ as 
\begin{multline}
    d(V_i, V_j) = \sum\limits_{n=1}^{n=12}\sqrt{(x_{i,n} - x_{j,n})^2 + (y_{i,n} - y_{j,n})^2} \\ + k_a | a_i - a_j | + k_v | v_i - v_j | + k_h |h_i - h_j|
\end{multline}

We use $k$ as a scaling parameter to weigh different aspects of the vehicle state in relation to the position error. In our experiments, we use $k_h = 1$ since heading change rates tend to range from -0.5 to 0.5, $k_v = 1/40$, with velocity ranging from 0 to 20, and $k_a = 1/20$, with acceleration ranging from -5 to 5. These values can be further changed to reflect the importance of different parts of the trajectory-state for clustering applications to identify different types of trajectory corner cases \cite{rosch2022space}. 

\begin{figure*}
    \centering
    \includegraphics[trim={1cm 0 1.5cm 1.5cm},clip, width=.16\textwidth]{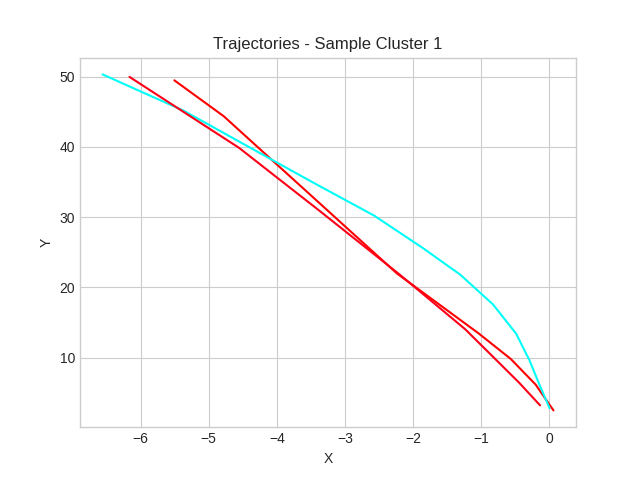}
    \includegraphics[trim={1cm 0 1.5cm 1.5cm},clip,width=.16\textwidth]{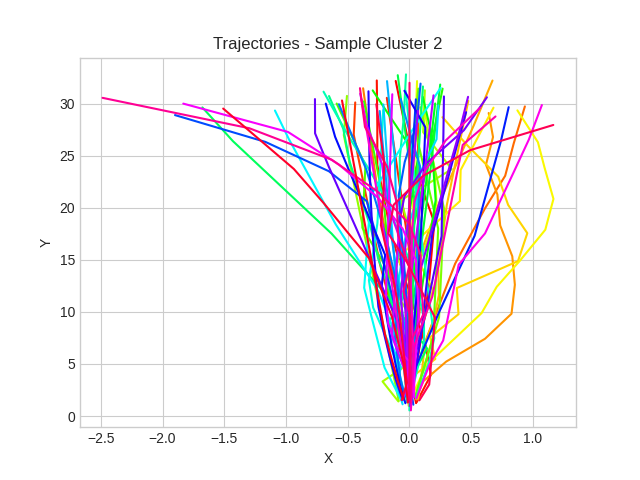}
    \includegraphics[trim={1cm 0 1.5cm 1.5cm},clip,width=.16\textwidth]{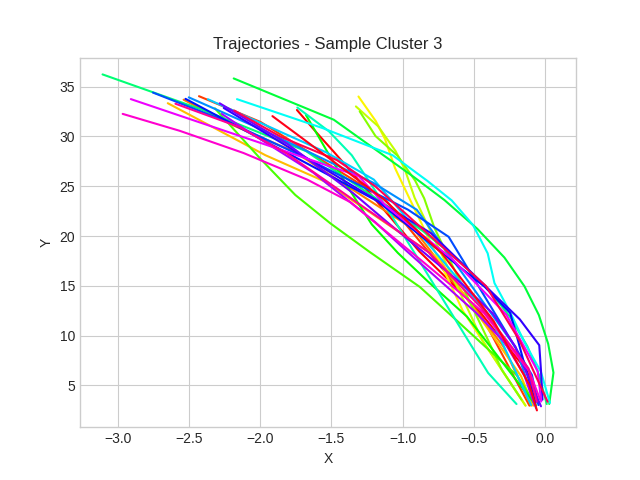}
    \includegraphics[trim={1cm 0 1.5cm 1.5cm},clip,width=.16\textwidth]{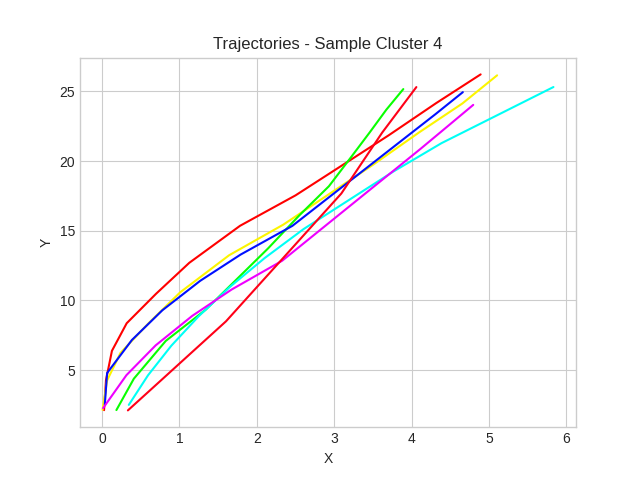}
    \includegraphics[trim={1cm 0 1.5cm 1.5cm},clip,width=.16\textwidth]{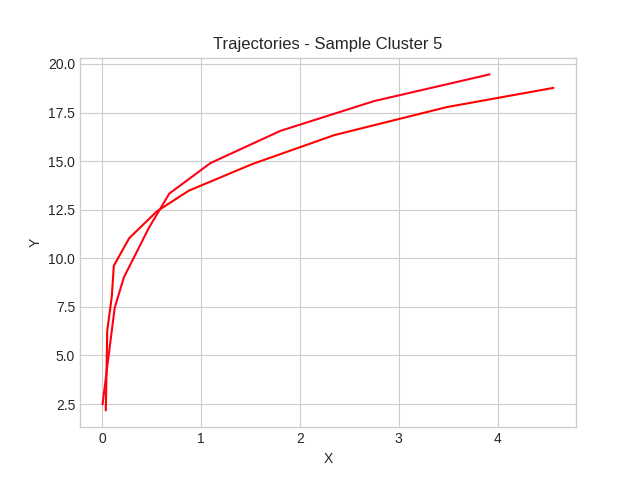}
    \includegraphics[trim={1cm 0 1.5cm 1.5cm},clip,width=.16\textwidth]{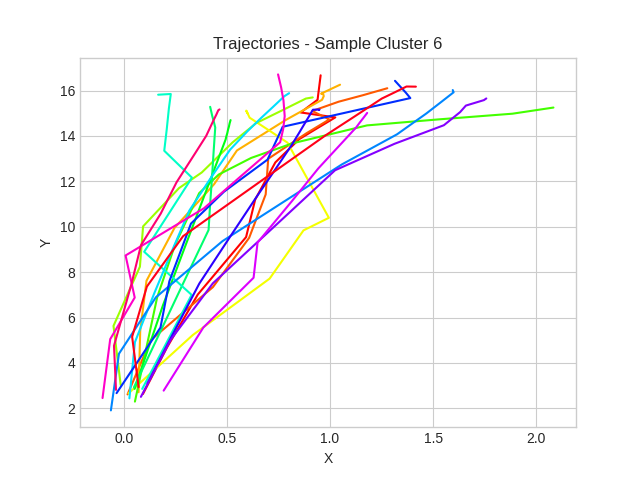}
    \includegraphics[trim={1cm 0 1.5cm 1.5cm},clip,width=.16\textwidth]{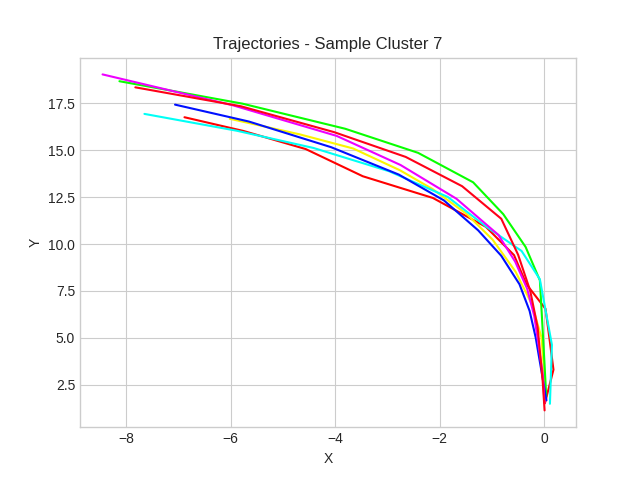}
    \includegraphics[trim={1cm 0 1.5cm 1.5cm},clip,width=.16\textwidth]{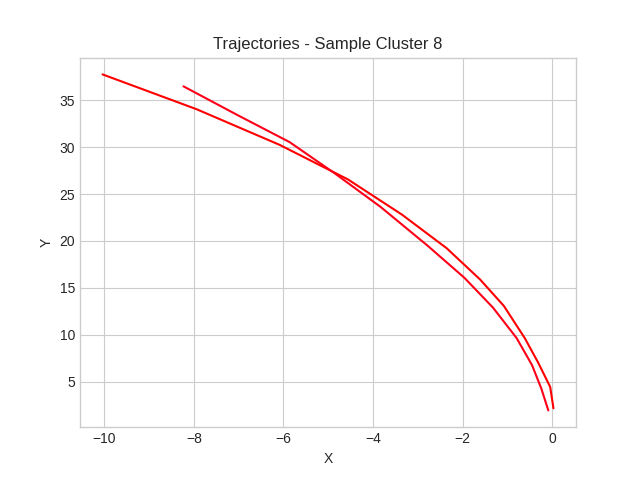}
    \includegraphics[trim={1cm 0 1.5cm 1.5cm},clip,width=.16\textwidth]{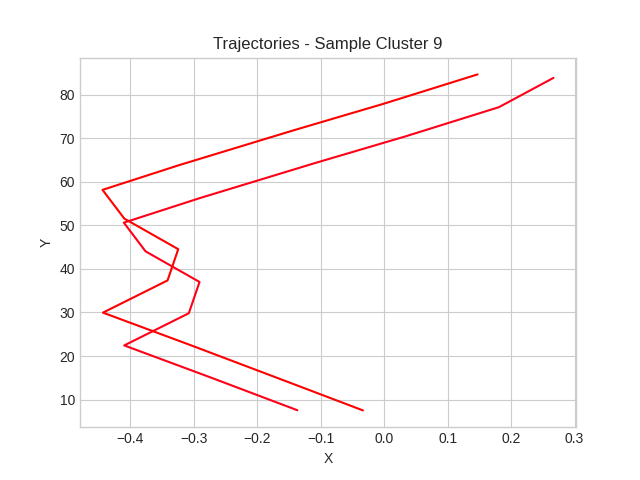}
    \includegraphics[trim={1cm 0 1.5cm 1.5cm},clip,width=.16\textwidth]{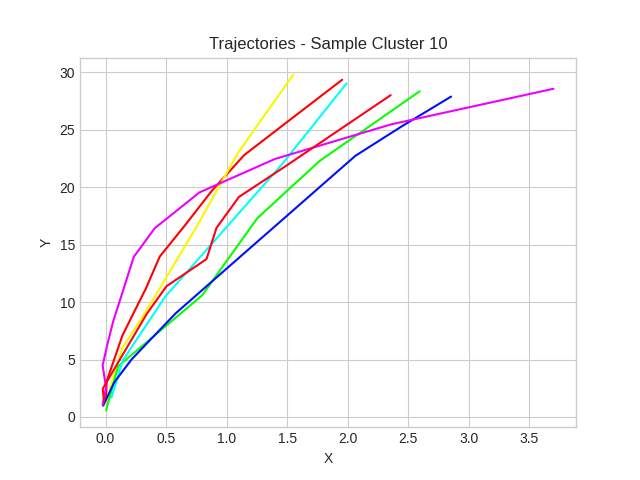}
    \includegraphics[trim={1cm 0 1.5cm 1.5cm},clip,width=.16\textwidth]{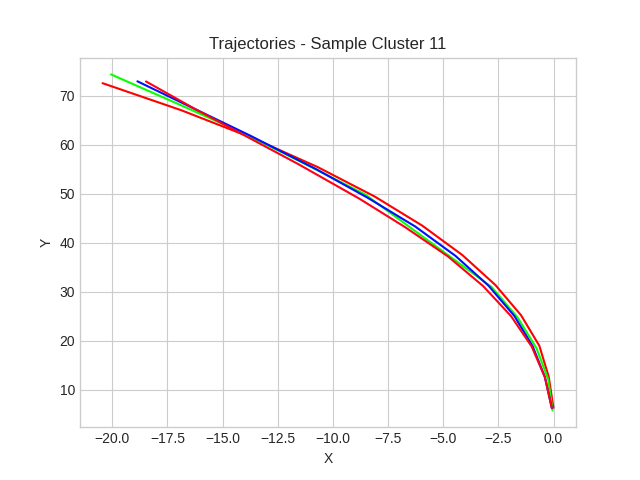}
    \includegraphics[trim={1cm 0 1.5cm 1.5cm},clip,width=.16\textwidth]{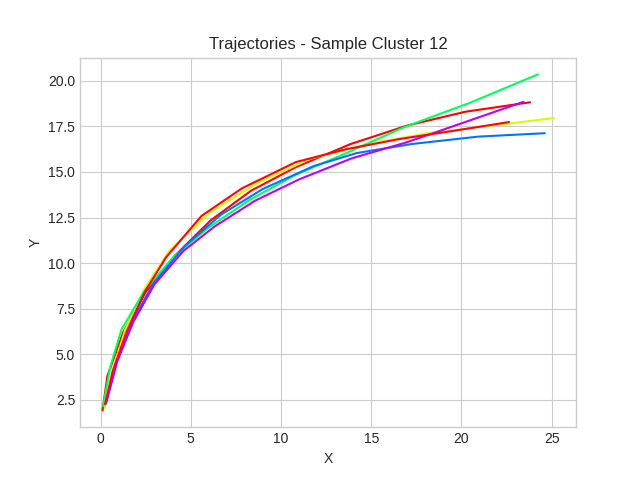}
    \caption{We randomly select 12 clusters, formed using our distance measurement over trajectory-states (which include trajectory coordinates and vehicle dynamics). Comparing across the selected clusters, clear patterns emerge even over the 2D coordinates alone (visualized), showing the effectiveness of grouping like-trajectories.}
    \label{tumsamples}
\end{figure*}

We apply the hierarchical clustering algorithm \cite{mullner2011modern, bar2001fast}, using the average distance of all points in a cluster in re-assigning cluster distances when constructing the dendrogram (i.e. unweighted pair group method with arithmetic mean). A threshold $\tau$ is applied to estimate the flat clusters, such that the cophenetic distance between any pair within one of the flat clusters is no greater than $\tau$. We use $\tau = 10$ in our experiments. Results of clustering are illustrated in Figure \ref{tumsamples}, showing just 12 of the 3,267 clusters formed in application of this algorithm to our experimental dataset, sampled at random. We also randomly sample a subset of 20 of the ``novel" trajectories (i.e. those which did not belong to a cluster), shown in Figure \ref{fig:enter-label} and illustrating that our trajectory-state distance measurement is effective in grouping like-trajectories and separating unique trajectories. 

\begin{algorithm}
\caption{Novelty-Sensitive Active Learning Round}
\label{alg:novelty_sensitive}
\SetKwInOut{Require}{Require}
\Require{$\alpha$, $\beta$, initial training pool $T_l$, unlabeled pool $T_u$, and data budget $B$}
\SetKwData{novel}{novel samples}
\SetKwData{familiar}{familiar samples}
\SetKwData{cluster}{cluster}
\SetKwData{pool}{training pool}
\SetKwData{unclustered}{unclustered set}
\SetKwData{overlap}{overlap}
\SetKwData{trajectories}{trajectories}
\SetKwData{samples}{samples}
\SetKwData{membership}{membership}
\SetKwData{annotations}{annotations}
\SetKwData{data}{data}
\SetKwData{budget}{data budget}
\SetKwData{percent}{\%}
\SetKwData{total}{total}
\SetKwFunction{Cluster}{Cluster}
\SetKwFunction{Select}{RandomSelect}
\SetKwFunction{Annotate}{Annotate}

$\novel \leftarrow {\emptyset}$ \;
$\familiar \leftarrow {\emptyset}$ \;
\Cluster{trajectorystates $v_i \in T_l \cup T_u$} \;
\While{$\|\novel\| < \alpha \times B$}{
    $C_n \leftarrow c_i \| \forall v_i \in c_i, v_i \notin T_l$\;
    $V_n \leftarrow v_i \| \forall c_i, v_i \notin c_i$\; 
    $S_n \leftarrow$ \Select{$s_i \in C_n \cup V_n$} \;
    \If{$S_n \in C_n$}{
        \For{$i = 1$ \KwTo $\beta \times \|S_n\|$} {
        $\novel$ += \Select{$v_i \in S_n$} \;
        }
    }
    \Else {
        $\novel += S_n$ \;
    }
}
\Annotate{\novel}\;
$T_l$ += \novel \;

\While{$\| \familiar \| < (1 - \alpha) \times B$}{
   $ C_t \leftarrow c_i \| \forall v_i \in c_i, \exists v_i \in T_l$\;
    \For{$i = 1$ \KwTo $\beta \times \|C_t\|$} {
        $\familiar += $ \Select{$s_i \in C_t$}\;
        }
    }
\Annotate{\familiar}\;
$T_l += \familiar$ \;
\end{algorithm}

\subsection{Active Learning Algorithm}

With the clusters available, it now becomes possible to use this unsupervised information within an active learning algorithm to enhance the learner's ability to intelligently acquire new data for annotation. This method does not require any measure of uncertainty from the learning model, but does require awareness of the currently-annotated training pool, as the membership of annotated trajectories within a cluster may disqualify that cluster from being acquired as `novel'. 

We present our algorithm in Algorithm \ref{alg:novelty_sensitive}. In summary, two parameters are used to define the included breadth (amount of `novel' samples added) and depth (how much of a cluster to be added) of the acquisition and annotation of new samples to the training pool. Up to these limits, clusters which are unrepresented in the training data, or singleton unclustered unique instances, can be drawn and added to the training pool, until the data budget is filled. In the case that no further sampling of the desired type is possible (e.g. there are no unvisited clusters or unique samples remaining), samples are drawn at random from the entire unlabeled pool.

\begin{figure}
    \centering
    \includegraphics[width=.4\textwidth]{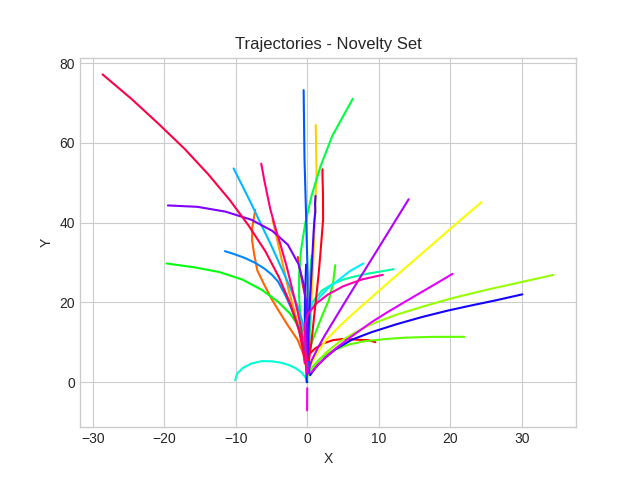}
    \caption{Many trajectory-states remain unclustered due to sufficient distance from all nearest trajectory-state clusters. We randomly sample just 20 of these unmatched trajectory-states, visualizing the 2D path coordinates and illustrating the diversity of behaviors found to be unique within the dataset.}
    \label{fig:enter-label}
\end{figure}

\section{Evaluation: Real-World Data and Experimental Design}

\subsection{Datasets}

We perform our experiments on the nuScenes dataset, using a subset of the public ``train" training split for training and another for validation, and the public ``train\_val" subset for testing. The nuScenes dataset contains 850 driving scenes for training and 150 for evaluation, divided into instances with 2 seconds of past history to be used in predicting 6 seconds into the future. Information on ego and surround vehicle state are available, as well as map structure (including lanes and intersections).  

\subsection{Experimental Design}

\begin{figure*}
    \centering
    \includegraphics[trim={2.95cm 0 5.6cm 1.5cm},clip, width=.186\textwidth]{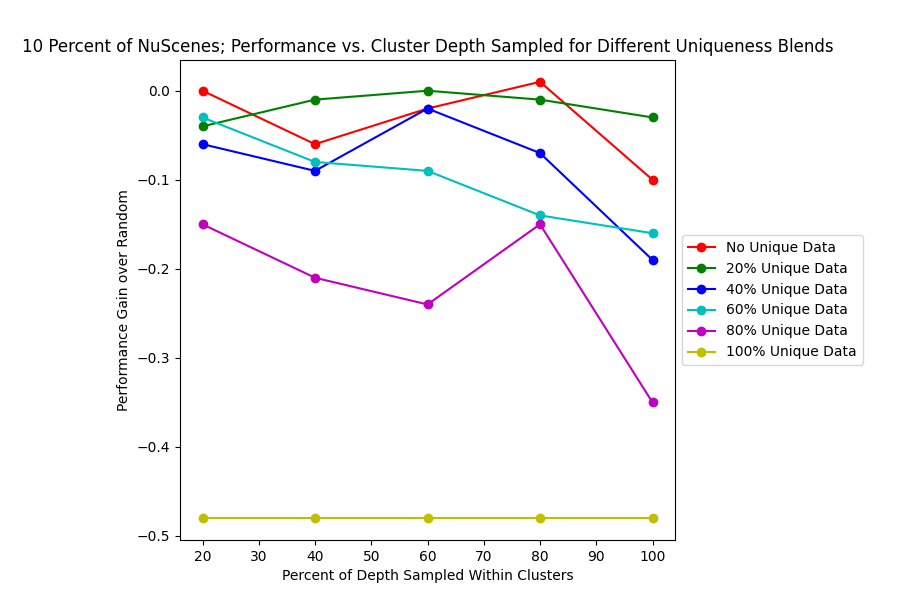}
    \includegraphics[trim={3.1cm 0 5.6cm 1.5cm},clip,width=.183\textwidth]{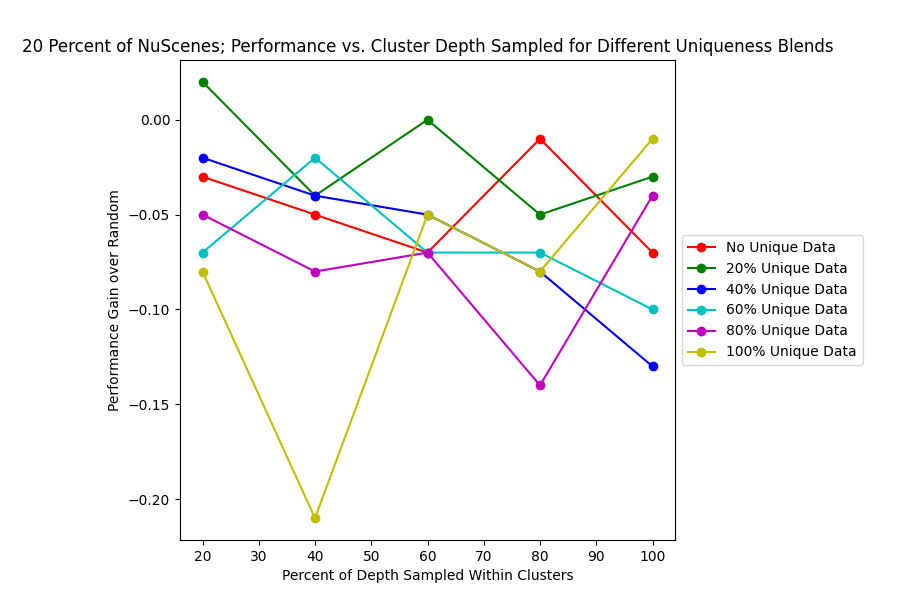}
    \includegraphics[trim={3.1cm  0 5.6cm 1.5cm},clip,width=.183\textwidth]{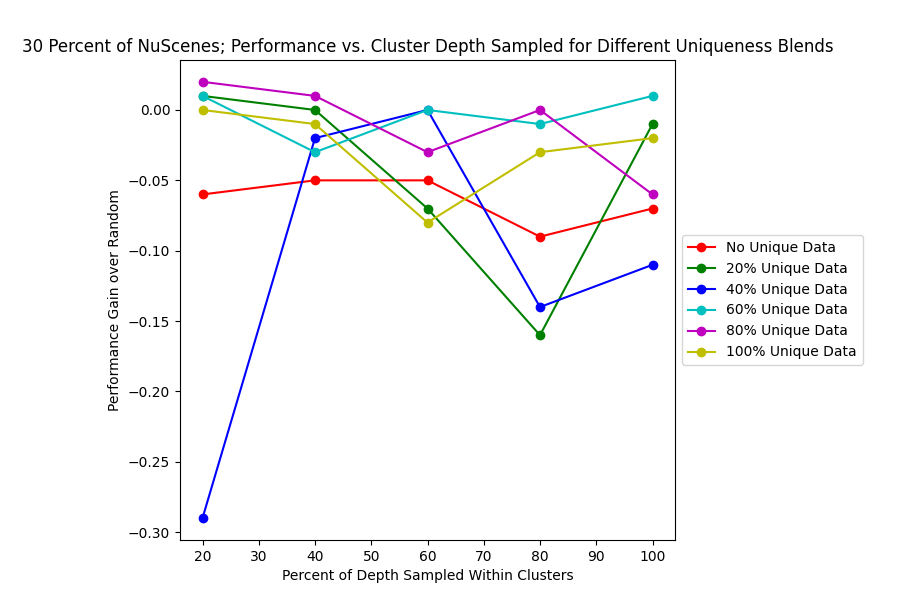}
    \includegraphics[trim={3.1cm  0 5.6cm 1.5cm},clip,width=.183\textwidth]{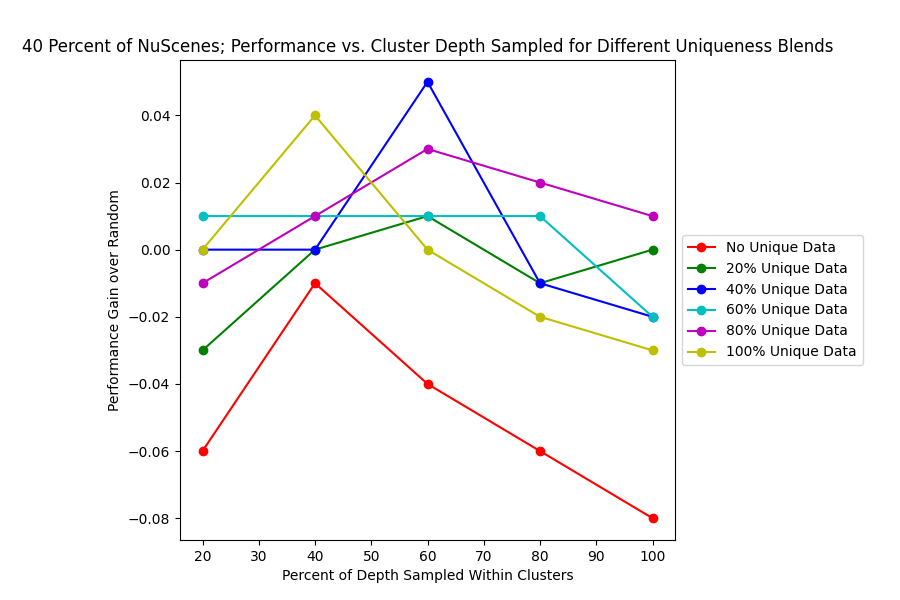}
    \includegraphics[trim={3cm  0 1.25cm 1.5cm},clip,width=.24\textwidth]{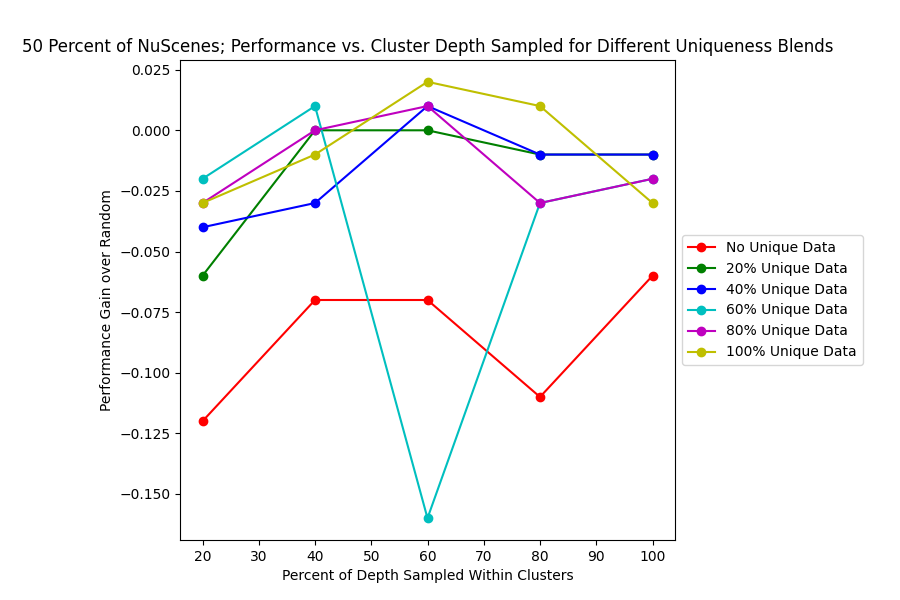}
    \caption{These five graphs represent the minimum average displacement error metric ($mADE_5$) performance of various parameterizations of the active learning strategy over a random baseline, considering the 5 most likely trajectory predictions from the model. Positive numbers indicate improvement over random. From left to right, each graph has a different training pool size, with the amount of data in the training pool increases from 10\% to 50\% of nuScenes (in 10\% increments). The y-axis represents improvement over random, while the x-axis represents the allowable ``depth" into a cluster that the algorithm samples. Each color line represents a different proportion of unique (novel, diverse) data, versus resampling data which is similar (typical) to data which already exists in the training pool. The point that we seek to highlight is the change in position of the yellow line (all novel data) and the red line (all typical data). We see that as the annotation budget or training pool size increases, these two trends effectively switch roles in over- (or under-) performing relative to the random baseline. This pattern matches the findings of Guy et al. in image classification tasks, providing evidence for the presence of the active learning phase transition within the trajectory prediction task - and, within the bounds of the nuScenes dataset size. Sampling typical data helps in overcoming a cold start, while novel data should be sampled in higher proportion as the training pool grows.}
    \label{samples}
\end{figure*}

\begin{figure*}
    \centering
    \includegraphics[trim={2.75cm 0 5.6cm 1.5cm},clip, width=.186\textwidth]{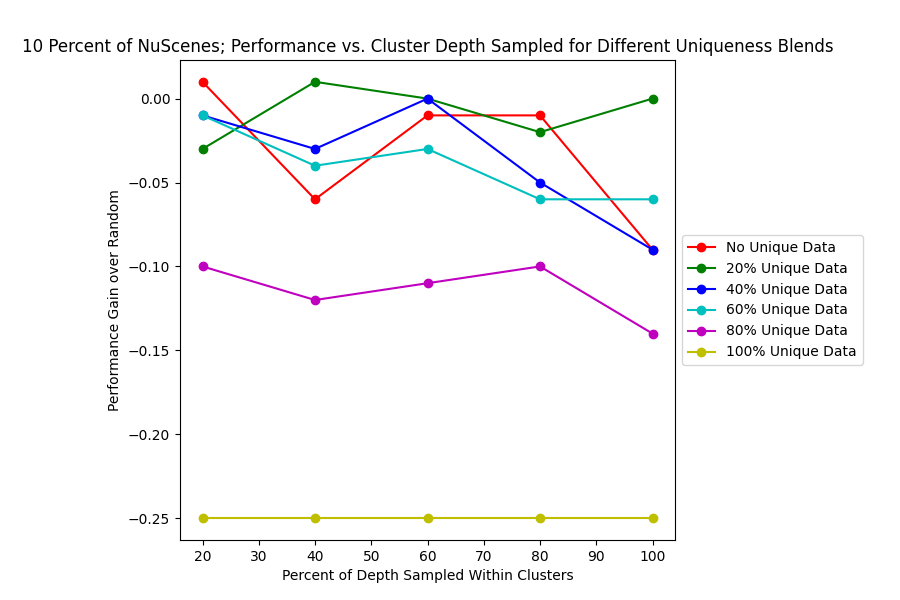}
    \includegraphics[trim={3.1cm 0 5.6cm 1.5cm},clip,width=.183\textwidth]{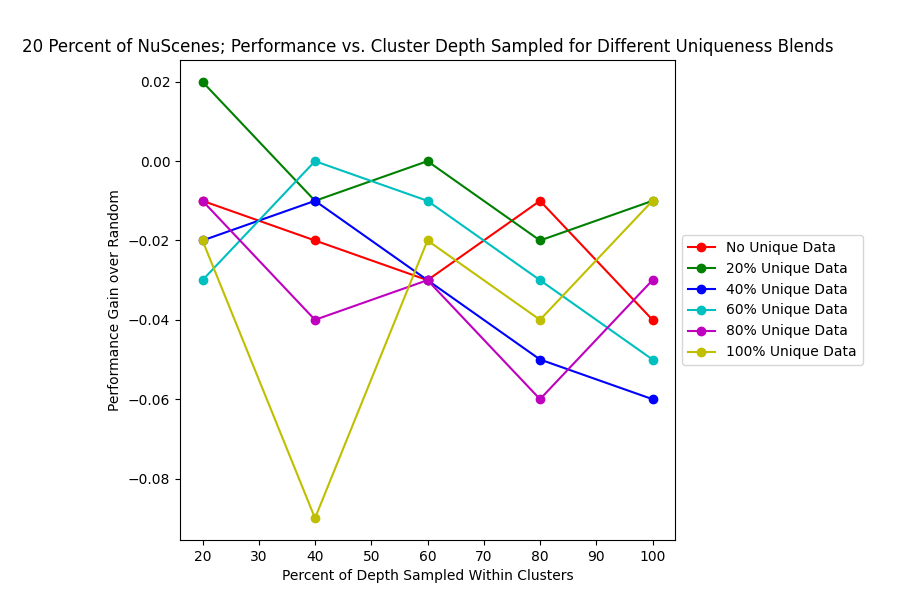}
    \includegraphics[trim={3.1cm  0 5.6cm 1.5cm},clip,width=.183\textwidth]{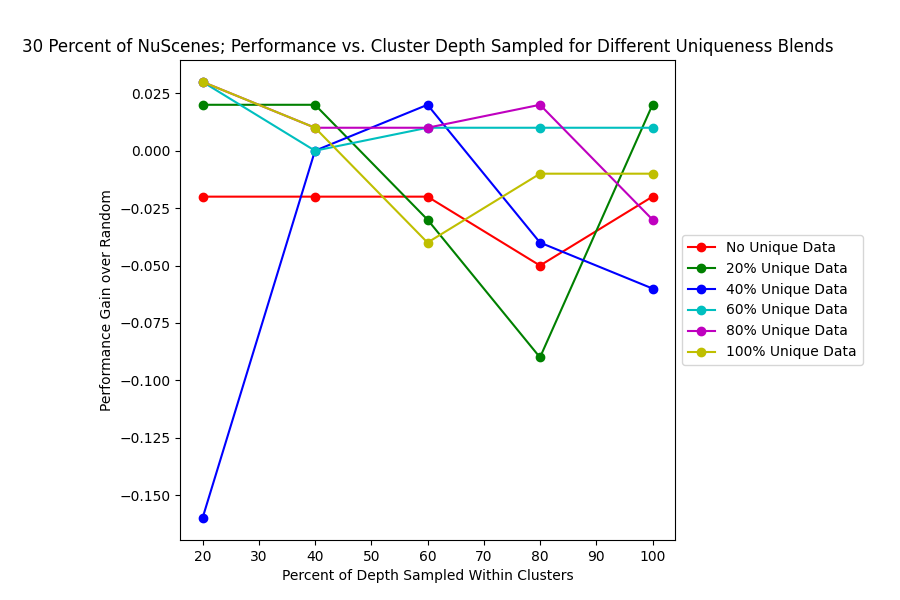}
    \includegraphics[trim={3.1cm  0 5.6cm 1.5cm},clip,width=.183\textwidth]{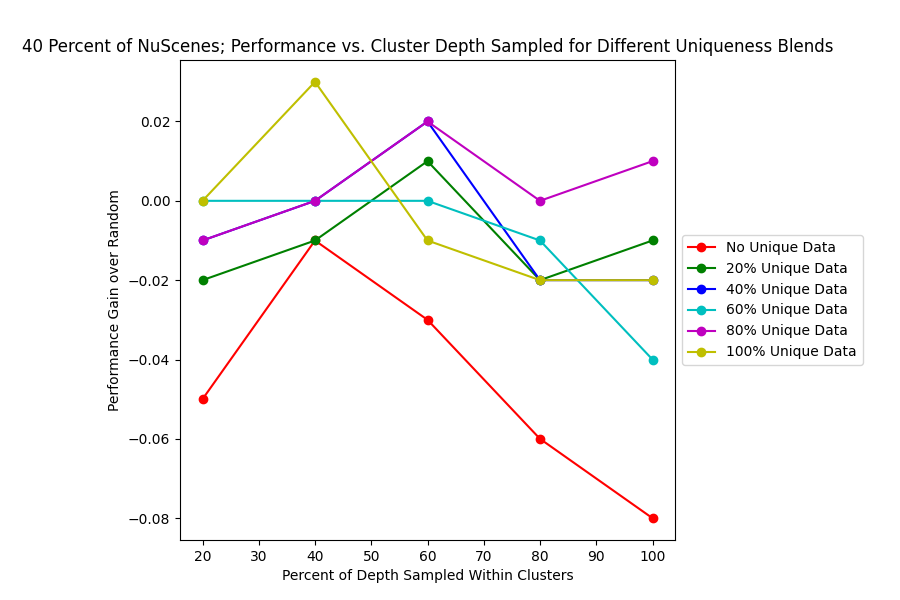}
    \includegraphics[trim={3cm  0 1.25cm 1.5cm},clip,width=.24\textwidth]{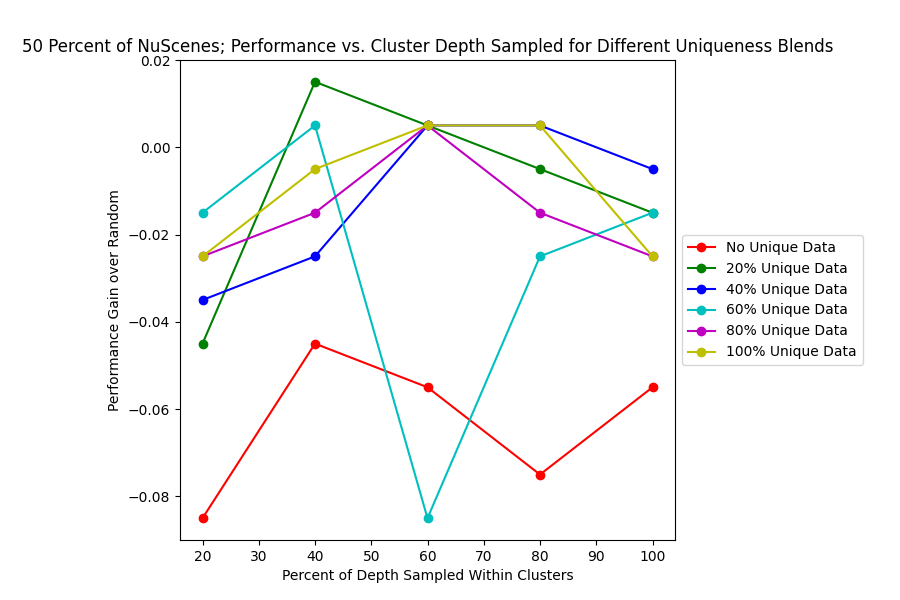}
    \caption{These five graphs represent the minimum average displacement error metric ($mADE_{10}$) performance of various parameterizations of the active learning strategy over a random baseline, considering the 10 most likely trajectory predictions from the model. Positive numbers indicate improvement over random. From left to right, each graph has a different training pool size, with the amount of data in the training pool increases from 10\% to 50\% of nuScenes (in 10\% increments). The y-axis represents improvement over random, while the x-axis represents the allowable ``depth" into a cluster that the algorithm samples. Each color line represents a different proportion of unique (novel, diverse) data, versus resampling data which is similar (typical) to data which already exists in the training pool. We observe the same pattern as noted in the graphs of $mADE_5$, in the transposition of performance of the strategy which samples novel data and the strategy which samples typical data.}
    \label{samples2}
\end{figure*}

We sweep through $\alpha$ and $\beta$ parameters in 20\% increments, beginning at $\alpha = 0$ (no novel data) and $\beta = .2$ (maximum number of samples from a given cluster is 20\% of the cluster size). We repeat this sweep on 5 training volumes: 10\% of the dataset through 50\% of the dataset, in 10\% increments. Results are illustrated in Figures \ref{samples} and \ref{samples2} and summarized in Table \ref{tab:nuscenesprogress_random_entropy}. 

We use the Prediction via Graph-based Policy (PGP) model \cite{deo2022multimodal} as the trajectory prediction model for training in the active learning framework. PGP learns discrete policies, exploring lane graph goals and waypoints with consideration for both lateral variability (lanekeeping, turning) and longitudinal variability (acceleration). PGP is one of the top models in trajectory prediction at the time of this writing, with top-3 performance on minimum average displacement and miss rate metrics of the nuScenes leaderboard, but regardless, the methods described in this paper are applicable to any machine learning system for trajectory prediction.

\section{Analysis of Results}

We present the results of our experiment in Table \ref{tab:nuscenesprogress_random_entropy}. This table presents performance of the best-performing active learning strategy in comparison to a random baseline on two common trajectory prediction metrics, the minimum average displacement error over the five highest-probability trajectories ($minADE_5$), and the same error over the ten highest-probability trajectories ($minADE_{10}$). We measure these values at five different data pool sizes, up to 50\% of the complete nuScenes training set. We also compare to the performance reported by \cite{deo2022multimodal} in the original PGP paper. Our diversity-driven active learning methods show consistent performance gains over random sampling, surpassing or equivalent at all data pool sizes, and even reaching (or surpassing) performance on the full dataset at just a fraction of the training pool size.

These methods do rely on selection of $\alpha, \beta$ parameters; in our table, we have the experimental luxury of providing the optimal values, but in practice, this would require some assessment of whether a model for a particular task has passed the point of inflection for active learning ``phase"; that is, whether or not it is more beneficial to sample \textit{typical} data or \textit{novel} data. The trend in our table, and in the associated figures, is still apparent: it is beneficial to sample typicality at the beginning, to address the ``cold-start problem", and as the data budget increases, begin introducing more-and-more novelty. We see at the 20\% budget, we accept 20\% novelty (one increment up from the initial 0\%), and in the higher budget sizes of 30-50\%, we begin finding the higher novelty $\alpha$ values to be optimal, making the case for some form of ``novelty scheduling" to be integrated into learning systems as a means of active learning. 

\begin{table*}[]
    \centering
        \caption{Active learning strategy performance compared to random baseline at five data pool sizes.}
    \begin{adjustbox}{width=.75\textwidth}
    \begin{tabular}{c||c|c|c||c|c|c}
        \hline 
        Labeled Pool & \multicolumn{3}{c||}{$mADE_5$} & \multicolumn{3}{c}{$mADE_{10}$} \\
        \hline
         \% & Random & Active & $\alpha, \beta$ & Random & Active & $\alpha, \beta$ \\
        \hline \hline 
        \rule{0pt}{2ex}  10\% & 1.59 & \textbf{1.58} (\scriptsize{\color{ForestGreen}{$-$0.01}}) &  (0\%, 20\%) & 1.18 & \textbf{1.17} (\scriptsize{\color{ForestGreen}{$-$0.01}}) & (0\%, 60\%) \\
        \hline
        \rule{0pt}{2ex}  20\% & 1.44 & \textbf{1.42} (\scriptsize{\color{ForestGreen}{$-$0.02}}) &  (20\%, 20\%) & 1.08 & \textbf{1.06} (\scriptsize{\color{ForestGreen}{$-$0.02}}) &  (20\%, 20\%)  \\
        \hline
         \rule{0pt}{2ex}  30\% & 1.37 & \textbf{1.35} (\scriptsize{\color{ForestGreen}{$-$0.02}}) &  (80\%, 20\%) & 1.04 & \textbf{1.01} (\scriptsize{\color{ForestGreen}{$-$0.03}}) & (60\%, 20\%)  \\
        \hline
        \rule{0pt}{2ex} 40\% & 1.35 & \textbf{1.30} (\scriptsize{\color{ForestGreen}{$-$0.05}}) & (40\%, 60\%) & 1.00 & \textbf{0.97} (\scriptsize{\color{ForestGreen}{$-$0.03}}) & (100\%, 40\%)  \\
        \hline
        \rule{0pt}{2ex} 50\% & 1.31 & \textbf{1.29} (\scriptsize{\color{ForestGreen}{$-$0.02}}) & (100\%, 40\%) & 0.97 & \textbf{0.96} (\scriptsize{\color{ForestGreen}{$-$0.01}}) & (20\%, 40\%) \\
        \hline
        \hline
         SoA 100\% \cite{deo2022multimodal} &\multicolumn{3}{c||}{\textbf{1.30}} & \multicolumn{3}{c}{\textbf{1.00}}\\
        \hline
    \end{tabular}
    \end{adjustbox}

    \label{tab:nuscenesprogress_random_entropy}
\end{table*}

\begin{figure}
    \centering
    \includegraphics[width=.49\textwidth]{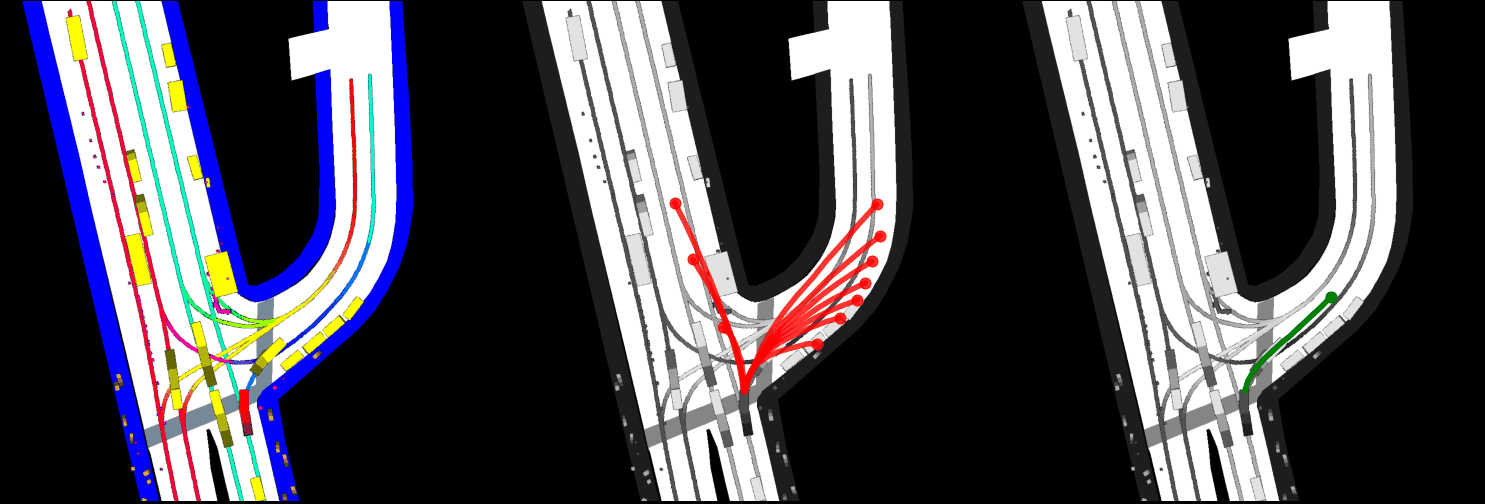}
        \includegraphics[width=.49\textwidth]{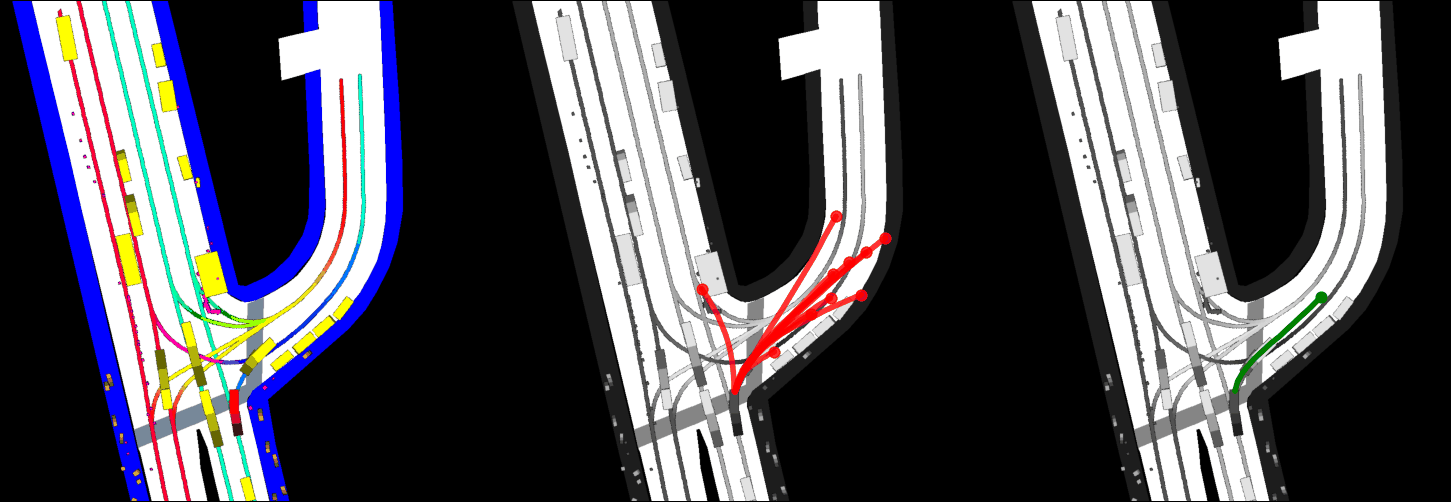}
    \includegraphics[width=.49\textwidth]{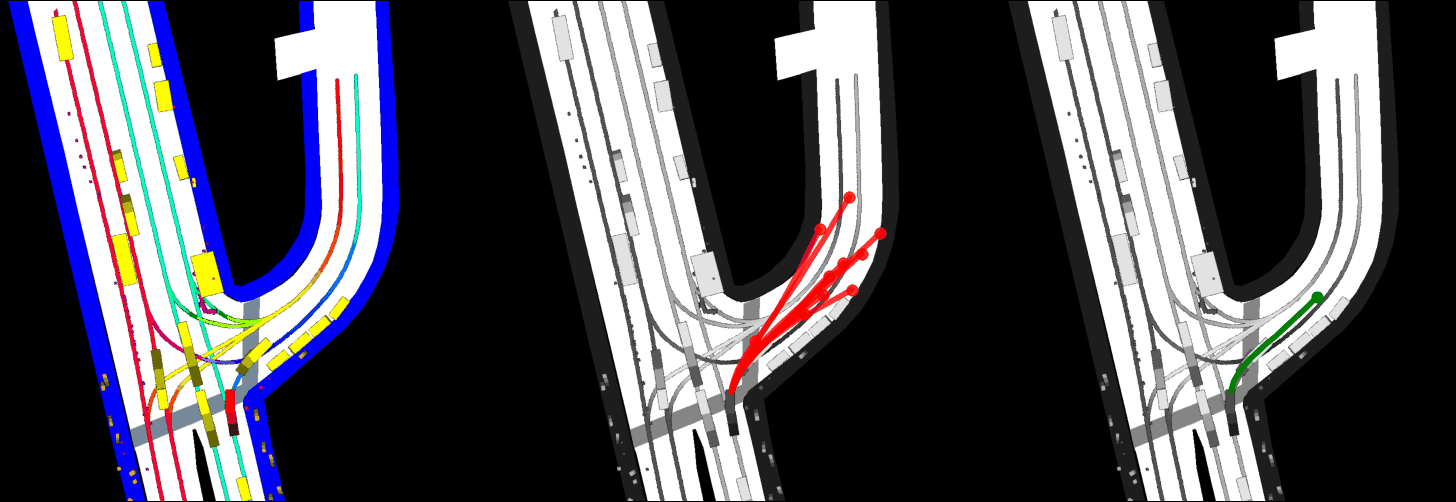}
    \caption{From left to right: map input, predicted trajectories, and ground truth. From top to bottom, results from models trained on: 10\% training data, 50\% training data randomly selected, and 50\% training data selected using our active learning algorithm. The 10\% data model shows a large spread of possible trajectories, with little scene conformity. Though the scene conformity improves at 50\% data, with active learning, the trajectories adapt even better to the lane contours of the scene.}
    \label{fig:ex1}
\end{figure}

\begin{figure}
    \centering
    \includegraphics[width=.49\textwidth]{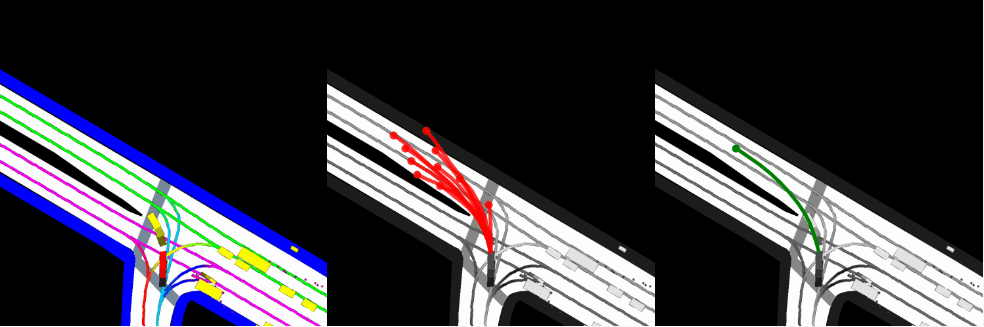}
        \includegraphics[width=.49\textwidth]{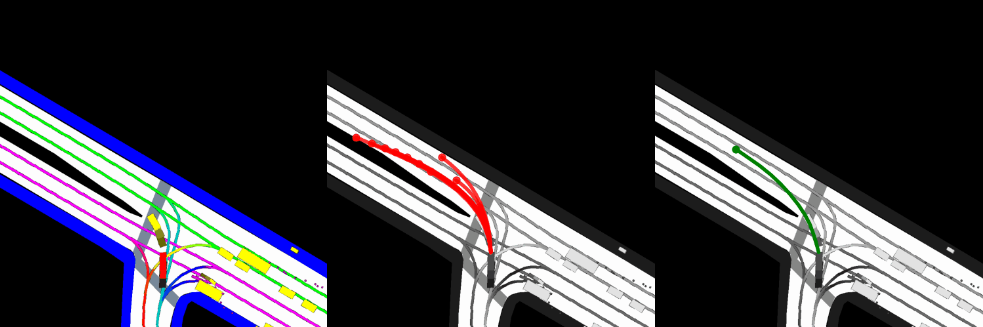}
    \includegraphics[width=.49\textwidth]{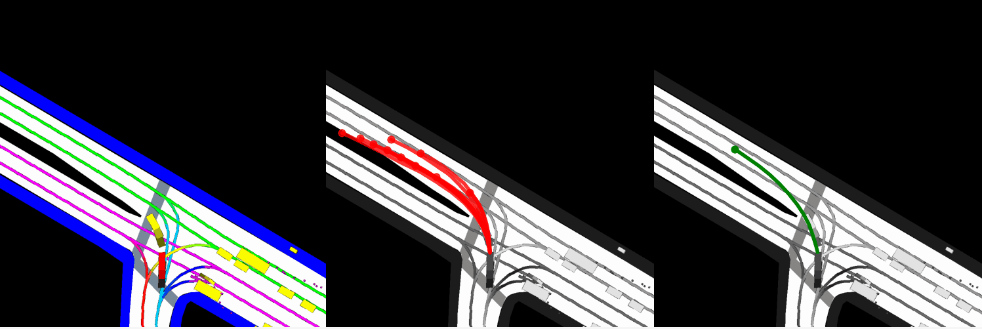}
    \caption{As in the previous example, though the scene conformity improves at 50\% data, with active learning, the trajectories snap more closely to the lanes in the freeway, while maintaining the multimodal options appropriate for the driver's choice in the scene.}
    \label{fig:ex2}
\end{figure}

\begin{figure}
    \centering
    \includegraphics[width=.49\textwidth]{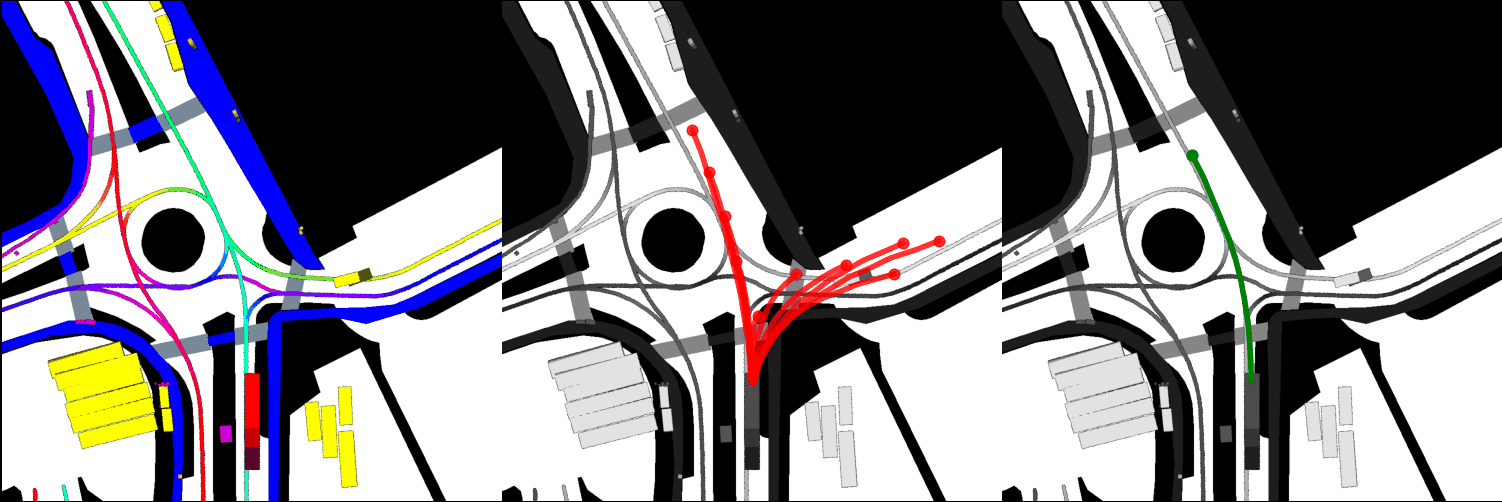}
        \includegraphics[width=.49\textwidth]{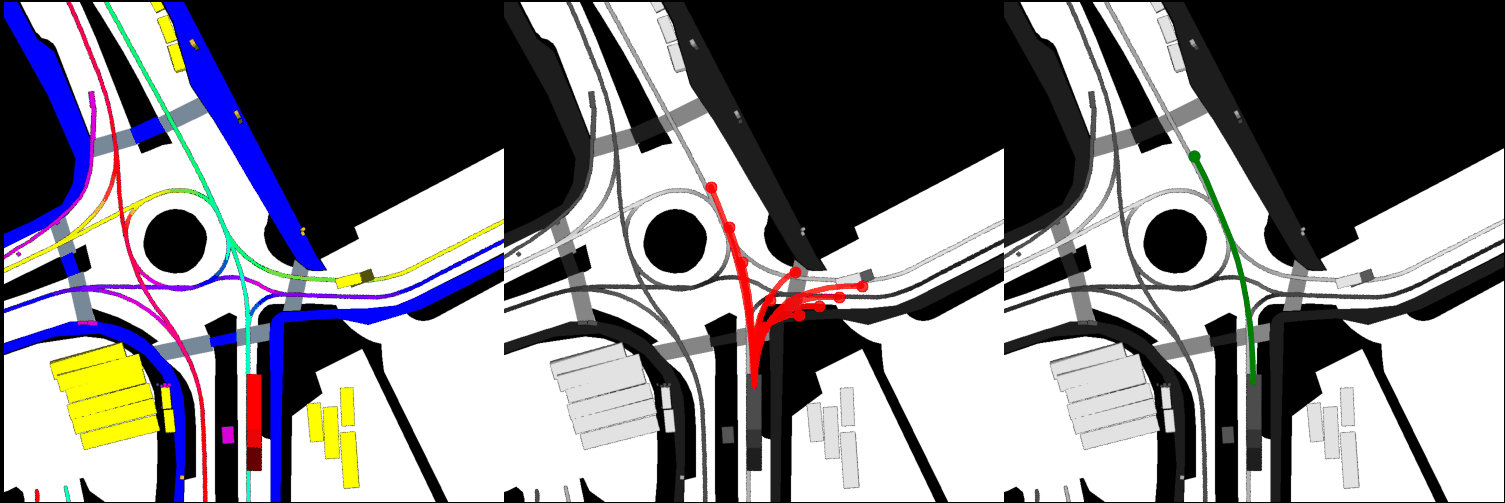}
    \includegraphics[width=.49\textwidth]{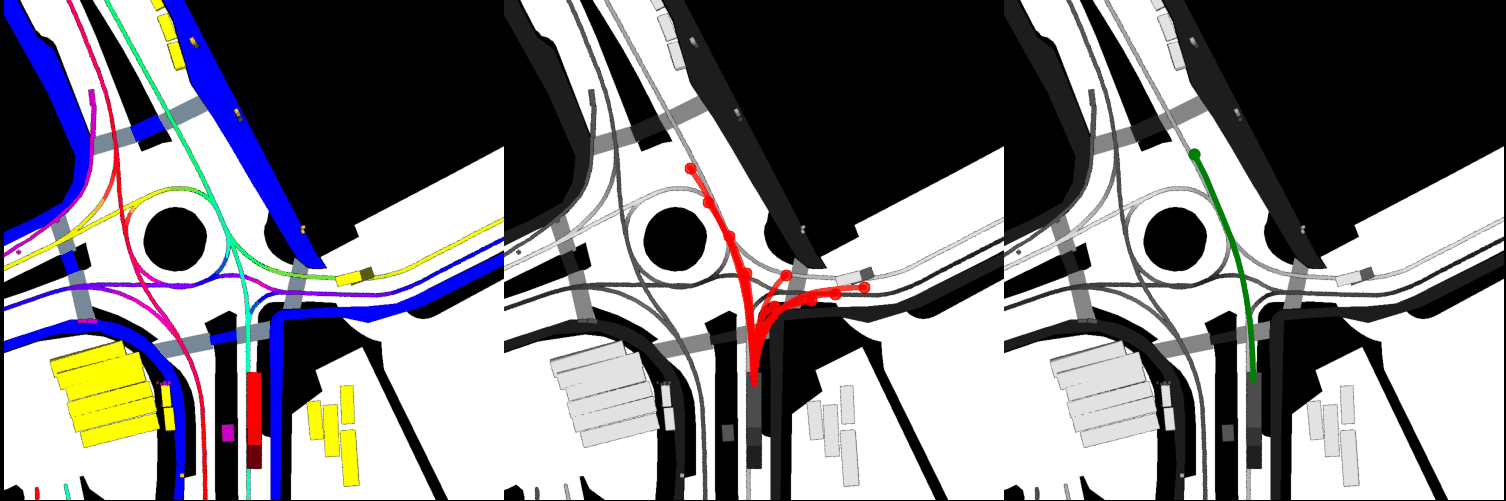}
    \caption{At the roundabout, many trajectories proposed by the 10\% training data model are non-compliant, and while the 50\% randomly-selected training data model shows a better conformity to the two possible modes, the `right turn' mode still has four very distinct (incorrect) variants. At 50\% active-learning-selected training data, these four variants collapse to one (scene-appropriate) turn.}
    \label{fig:ex3}
\end{figure}

\begin{figure}
    \centering
    \includegraphics[width=.49\textwidth]{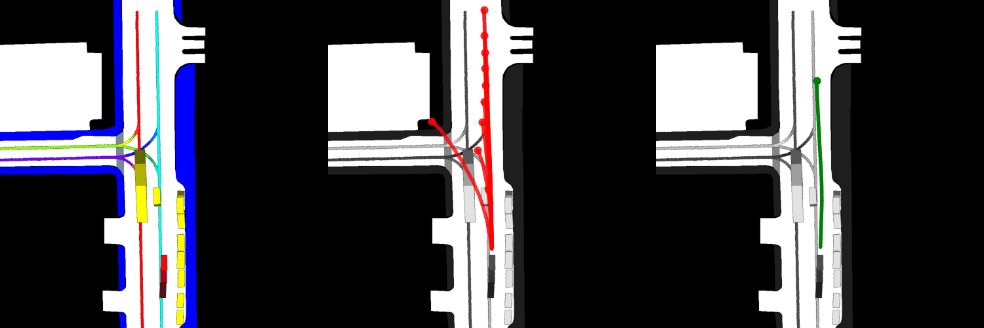}
        \includegraphics[width=.49\textwidth]{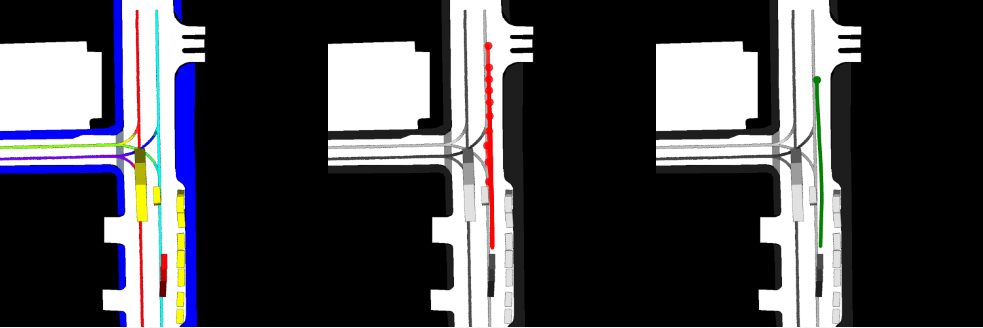}
    \includegraphics[width=.49\textwidth]{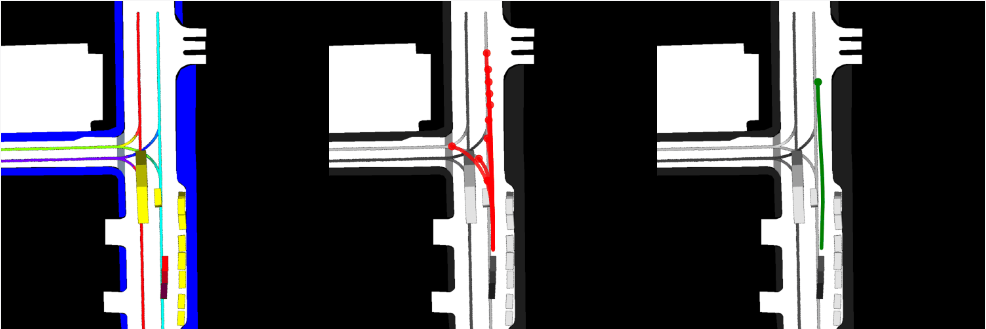}
    \caption{Though the vehicle continues straight in this example, a good trajectory prediction model should maintain an awareness of the possibility of a left turn, as the future (in this case) is truly unpredictable since the driver has agency to elect to take the turn. The 10\% training data model has no real understanding of the map, and only produces some kinematically-possible modes which are inappropriate to the scene. The 50\% randomly selected training data model loses the mode for the left turn, while the active learning training method maintains both the mode \textit{and} strong lane adherence in all predictions.}
    \label{fig:ex4}
\end{figure}

Qualitative results are depicted in Figures \ref{fig:ex1}-\ref{fig:ex4}, with examples of model results from the ten percent data pool, fifty percent random data pool, and fifty percent active learning pool. Specific cases are discussed in the figure captions, and one pattern that emerges between examples is the pace of the model's learning of lane-conforming behavior and multimodality. In the ten percent data volume, the model begins with a wide spread of modal coverage, but rarely conforming to any of the scene input. Rather, the model very loosely approximates a variety of kinematically-feasible spreads, regardless of map state. At fifty percent data randomly selected, the model begins to converge toward lane configurations, but notably deviates from the lane centerlines at a greater rate than the fifty percent data efficiently sampled using our active learning algorithm. Further, as shown in Figure \ref{fig:ex4}, the active learning approach seems to show the same level of conformity even in its multi-trajectory predictions in the case of multiple possible futures. 

\section{Concluding Remarks}

In this research, we present a method by which information about a vehicle's trajectory and dynamic state, collectively referred to as a trajectory-state, can be clustered. We show the utility of these clusters as the drivers of a selection criteria in an active learning framework. However, this is not the only way to cluster such data, nor is this the only possible data which can be used in the clustering process; future research can iterate on these methods to further drive development of learning systems which select data at low cost to human annotators and intelligently guide data curation at scale. While we provide a selection based on ``novelty" or ``uniqueness" in this research, other measures, such as salience \cite{greer2022salience, greer2023salient, greer2023robust} or even language-based queries \cite{gopalkrishnan2024multi}, may also be highly informative to efficient and safe model learning \cite{ohn2020learning}. Beyond learning itself, such novelty-mining is also important in selection of data for system validation \cite{prakash2020exploring}.

Further, in this set of experiments, we apply the trajectory-state-informed active learning toward the task of trajectory prediction, but the utility should be further explored in additional autonomous driving tasks \cite{hekimoglu2023multi}. We make an argument at the beginning of the paper that one can infer much about the outside scene from the trajectory alone. While the outside scene is subsequently annotated and used for the trajectory prediction task we annotate, it would be interesting to see how well the trajectory informs active learning for the other relevant task of object detection, reinforcing the mutual information between the visual sensing of a scene and an agent's response trajectory to a scene (i.e. ``perception without vision"). 

In closing, we repeat that the proposed clustering and active learning algorithms are methods by which large-scale data systems can be more efficient without sacrificing performance on safety-critical predictive tasks. Data-driven methods show significant promise towards robust safety, but handling the long-tail nature of high-risk driving events requires intelligent approaches to collecting, curating, and annotating this valuable data.

\bibliographystyle{IEEEtran}
\bibliography{biblio.bib}

\begin{thebibliography}{10}
\providecommand{\url}[1]{#1}
\csname url@samestyle\endcsname
\providecommand{\newblock}{\relax}
\providecommand{\bibinfo}[2]{#2}
\providecommand{\BIBentrySTDinterwordspacing}{\spaceskip=0pt\relax}
\providecommand{\BIBentryALTinterwordstretchfactor}{4}
\providecommand{\BIBentryALTinterwordspacing}{\spaceskip=\fontdimen2\font plus
\BIBentryALTinterwordstretchfactor\fontdimen3\font minus \fontdimen4\font\relax}
\providecommand{\BIBforeignlanguage}[2]{{%
\expandafter\ifx\csname l@#1\endcsname\relax
\typeout{** WARNING: IEEEtran.bst: No hyphenation pattern has been}%
\typeout{** loaded for the language `#1'. Using the pattern for}%
\typeout{** the default language instead.}%
\else
\language=\csname l@#1\endcsname
\fi
#2}}
\providecommand{\BIBdecl}{\relax}
\BIBdecl

\bibitem{cui2019multimodal}
H.~Cui, V.~Radosavljevic, F.-C. Chou, T.-H. Lin, T.~Nguyen, T.-K. Huang, J.~Schneider, and N.~Djuric, ``Multimodal trajectory predictions for autonomous driving using deep convolutional networks,'' in \emph{2019 international conference on robotics and automation (icra)}.\hskip 1em plus 0.5em minus 0.4em\relax IEEE, 2019, pp. 2090--2096.

\bibitem{messaoud2021trajectory}
K.~Messaoud, N.~Deo, M.~M. Trivedi, and F.~Nashashibi, ``Trajectory prediction for autonomous driving based on multi-head attention with joint agent-map representation,'' in \emph{2021 IEEE Intelligent Vehicles Symposium (IV)}.\hskip 1em plus 0.5em minus 0.4em\relax IEEE, 2021, pp. 165--170.

\bibitem{kim2022diverse}
S.~Kim, H.~Jeon, J.~W. Choi, and D.~Kum, ``Diverse multiple trajectory prediction using a two-stage prediction network trained with lane loss,'' \emph{IEEE Robotics and Automation Letters}, vol.~8, no.~4, pp. 2038--2045, 2022.

\bibitem{greer2021trajectory}
R.~Greer, N.~Deo, and M.~Trivedi, ``Trajectory prediction in autonomous driving with a lane heading auxiliary loss,'' \emph{IEEE Robotics and Automation Letters}, vol.~6, no.~3, pp. 4907--4914, 2021.

\bibitem{deo2018convolutional}
N.~Deo and M.~M. Trivedi, ``Convolutional social pooling for vehicle trajectory prediction,'' in \emph{Proceedings of the IEEE conference on computer vision and pattern recognition workshops}, 2018, pp. 1468--1476.

\bibitem{deo2020trajectory}
------, ``Trajectory forecasts in unknown environments conditioned on grid-based plans,'' \emph{arXiv preprint arXiv:2001.00735}, 2020.

\bibitem{zimmer20193d}
W.~Zimmer, A.~Rangesh, and M.~Trivedi, ``3d bat: A semi-automatic, web-based 3d annotation toolbox for full-surround, multi-modal data streams,'' in \emph{2019 IEEE Intelligent Vehicles Symposium (IV)}.\hskip 1em plus 0.5em minus 0.4em\relax IEEE, 2019, pp. 1816--1821.

\bibitem{greer2024and}
R.~Greer, B.~Antoniussen, M.~V. Andersen, A.~M{\o}gelmose, and M.~M. Trivedi, ``The why, when, and how to use active learning in large-data-driven 3d object detection for safe autonomous driving: An empirical exploration,'' \emph{arXiv preprint arXiv:2401.16634}, 2024.

\bibitem{ruckin2024semi}
J.~R{\"u}ckin, F.~Magistri, C.~Stachniss, and M.~Popovi{\'c}, ``Semi-supervised active learning for semantic segmentation in unknown environments using informative path planning,'' \emph{IEEE Robotics and Automation Letters}, 2024.

\bibitem{almin2023navya3dseg}
A.~Almin, L.~Lemari{\'e}, A.~Duong, and B.~R. Kiran, ``Navya3dseg-navya 3d semantic segmentation dataset design \& split generation for autonomous vehicles,'' \emph{IEEE Robotics and Automation Letters}, 2023.

\bibitem{ghita2024activeanno3d}
A.~Ghita, B.~Antoniussen, W.~Zimmer, R.~Greer, C.~Cre{\ss}, A.~M{\o}gelmose, M.~M. Trivedi, and A.~C. Knoll, ``Activeanno3d--an active learning framework for multi-modal 3d object detection,'' \emph{arXiv preprint arXiv:2402.03235}, 2024.

\bibitem{sener2017active}
O.~Sener and S.~Savarese, ``Active learning for convolutional neural networks: A core-set approach,'' \emph{arXiv preprint arXiv:1708.00489}, 2017.

\bibitem{yang2015multi}
Y.~Yang, Z.~Ma, F.~Nie, X.~Chang, and A.~G. Hauptmann, ``Multi-class active learning by uncertainty sampling with diversity maximization,'' \emph{International Journal of Computer Vision}, vol. 113, pp. 113--127, 2015.

\bibitem{lu2024activead}
H.~Lu, X.~Jia, Y.~Xie, W.~Liao, X.~Yang, and J.~Yan, ``Activead: Planning-oriented active learning for end-to-end autonomous driving,'' \emph{arXiv preprint arXiv:2403.02877}, 2024.

\bibitem{sivaraman2010general}
S.~Sivaraman and M.~M. Trivedi, ``A general active-learning framework for on-road vehicle recognition and tracking,'' \emph{IEEE Transactions on intelligent transportation systems}, vol.~11, no.~2, pp. 267--276, 2010.

\bibitem{satzoda2015multipart}
R.~K. Satzoda and M.~M. Trivedi, ``Multipart vehicle detection using symmetry-derived analysis and active learning,'' \emph{IEEE Transactions on Intelligent Transportation Systems}, vol.~17, no.~4, pp. 926--937, 2015.

\bibitem{hacohen2022active}
G.~Hacohen, A.~Dekel, and D.~Weinshall, ``Active learning on a budget: Opposite strategies suit high and low budgets,'' \emph{arXiv preprint arXiv:2202.02794}, 2022.

\bibitem{caesar2020nuscenes}
H.~Caesar, V.~Bankiti, A.~H. Lang, S.~Vora, V.~E. Liong, Q.~Xu, A.~Krishnan, Y.~Pan, G.~Baldan, and O.~Beijbom, ``nuscenes: A multimodal dataset for autonomous driving,'' in \emph{Proceedings of the IEEE/CVF conference on computer vision and pattern recognition}, 2020, pp. 11\,621--11\,631.

\bibitem{rottmann2023automated}
M.~Rottmann and M.~Reese, ``Automated detection of label errors in semantic segmentation datasets via deep learning and uncertainty quantification,'' in \emph{Proceedings of the IEEE/CVF Winter Conference on Applications of Computer Vision}, 2023, pp. 3214--3223.

\bibitem{kulkarni2021create}
N.~Kulkarni, A.~Rangesh, J.~Buck, J.~Feltracco, M.~Trivedi, N.~Deo, R.~Greer, S.~Sarraf, and S.~Sathyanarayana, ``Create a large-scale video driving dataset with detailed attributes using amazon sagemaker ground truth,'' 2021.

\bibitem{zhu2019addressing}
Y.~Zhu, J.~Lin, S.~He, B.~Wang, Z.~Guan, H.~Liu, and D.~Cai, ``Addressing the item cold-start problem by attribute-driven active learning,'' \emph{IEEE Transactions on Knowledge and Data Engineering}, vol.~32, no.~4, pp. 631--644, 2019.

\bibitem{satzoda2014drive}
R.~K. Satzoda and M.~M. Trivedi, ``Drive analysis using vehicle dynamics and vision-based lane semantics,'' \emph{IEEE Transactions on Intelligent Transportation Systems}, vol.~16, no.~1, pp. 9--18, 2014.

\bibitem{rosch2022space}
K.~R{\"o}sch, F.~Heidecker, J.~Truetsch, K.~Kowol, C.~Schicktanz, M.~Bieshaare, B.~Sick, and C.~Stiller, ``Space, time, and interaction: A taxonomy of corner cases in trajectory datasets for automated driving,'' in \emph{2022 IEEE Symposium Series on Computational Intelligence (SSCI)}.\hskip 1em plus 0.5em minus 0.4em\relax IEEE, 2022, pp. 86--93.

\bibitem{mullner2011modern}
D.~M{\"u}llner, ``Modern hierarchical, agglomerative clustering algorithms,'' \emph{arXiv preprint arXiv:1109.2378}, 2011.

\bibitem{bar2001fast}
Z.~Bar-Joseph, D.~K. Gifford, and T.~S. Jaakkola, ``Fast optimal leaf ordering for hierarchical clustering,'' \emph{Bioinformatics}, vol.~17, no. suppl\_1, pp. S22--S29, 2001.

\bibitem{deo2022multimodal}
N.~Deo, E.~Wolff, and O.~Beijbom, ``Multimodal trajectory prediction conditioned on lane-graph traversals,'' in \emph{Conference on Robot Learning}.\hskip 1em plus 0.5em minus 0.4em\relax PMLR, 2022, pp. 203--212.

\bibitem{greer2022salience}
R.~Greer, J.~Isa, N.~Deo, A.~Rangesh, and M.~M. Trivedi, ``On salience-sensitive sign classification in autonomous vehicle path planning: Experimental explorations with a novel dataset,'' in \emph{Proceedings of the IEEE/CVF Winter Conference on Applications of Computer Vision}, 2022, pp. 636--644.

\bibitem{greer2023salient}
R.~Greer, A.~Gopalkrishnan, N.~Deo, A.~Rangesh, and M.~Trivedi, ``Salient sign detection in safe autonomous driving: Ai which reasons over full visual context,'' in \emph{27th International Technical Conference on the Enhanced Safety of Vehicles (ESV) National Highway Traffic Safety Administration}, no. 23-0333, 2023.

\bibitem{greer2023robust}
R.~Greer, A.~Gopalkrishnan, J.~Landgren, L.~Rakla, A.~Gopalan, and M.~Trivedi, ``Robust traffic light detection using salience-sensitive loss: Computational framework and evaluations,'' in \emph{2023 IEEE Intelligent Vehicles Symposium (IV)}, 2023, pp. 1--7.

\bibitem{gopalkrishnan2024multi}
A.~Gopalkrishnan, R.~Greer, and M.~Trivedi, ``Multi-frame, lightweight \& efficient vision-language models for question answering in autonomous driving,'' \emph{arXiv preprint arXiv:2403.19838}, 2024.

\bibitem{ohn2020learning}
E.~Ohn-Bar, A.~Prakash, A.~Behl, K.~Chitta, and A.~Geiger, ``Learning situational driving,'' in \emph{Proceedings of the IEEE/CVF Conference on Computer Vision and Pattern Recognition}, 2020, pp. 11\,296--11\,305.

\bibitem{prakash2020exploring}
A.~Prakash, A.~Behl, E.~Ohn-Bar, K.~Chitta, and A.~Geiger, ``Exploring data aggregation in policy learning for vision-based urban autonomous driving,'' in \emph{Proceedings of the IEEE/CVF Conference on Computer Vision and Pattern Recognition}, 2020, pp. 11\,763--11\,773.

\bibitem{hekimoglu2023multi}
A.~Hekimoglu, P.~Friedrich, W.~Zimmer, M.~Schmidt, A.~Marcos-Ramiro, and A.~Knoll, ``Multi-task consistency for active learning,'' in \emph{Proceedings of the IEEE/CVF International Conference on Computer Vision}, 2023, pp. 3415--3424.

\end{thebibliography}

\end{document}